\documentclass{article}

\PassOptionsToPackage{numbers}{natbib}

    \usepackage[preprint]{neurips_2025}

\usepackage{amsmath,amsfonts,bm}

\def\eqref#1{equation~\ref{#1}}

\def\1{\bm{1}}

\DeclareMathAlphabet{\mathsfit}{\encodingdefault}{\sfdefault}{m}{sl}
\SetMathAlphabet{\mathsfit}{bold}{\encodingdefault}{\sfdefault}{bx}{n}

\usepackage[utf8]{inputenc} 
\usepackage[T1]{fontenc}    
\usepackage{hyperref}       
\usepackage{url}            
\usepackage{booktabs}       
\usepackage{amsfonts}       
\usepackage{nicefrac}       
\usepackage{microtype}      
\usepackage{xcolor}         
\usepackage{graphicx}
\usepackage{booktabs}
\usepackage{url}
\usepackage{enumitem}
\usepackage[ruled,vlined]{algorithm2e}

\title{SITCOM: Scaling Inference Time Compute for Robotic Tasks}

\author{%
  Ayudh Saxena\thanks{Everyone Contributed Equally -- Ordering decided by rolling a dice} \quad
  Harsh Shah\footnotemark[1] \quad
  Sandeep Routray\footnotemark[1] \quad
  Rishi Rajesh Shah\footnotemark[1] \quad
  Esha Pahwa\footnotemark[1] \\
  Carnegie Mellon University \\
  \texttt{\{ayudhs, hshah2, sroutra2, rishisha, epahwa\}@cs.cmu.edu}
}

\title{SITCOM: Scaling Inference-Time COMpute for VLAs}

\date{}

\begin{document}
\maketitle

\begin{abstract}
Learning robust robotic control policies remains a major challenge due to the high cost of collecting labeled data, limited generalization to unseen environments, and difficulties in planning over long horizons. While Vision–Language–Action (VLA) models offer a promising solution by grounding natural language instructions into single-step control commands, they often lack mechanisms for lookahead and struggle with compounding errors in dynamic tasks. In this project, we introduce \emph{Scaling Inference-Time COMpute for VLAs} (\textbf{SITCOM}), a framework that augments any pretrained VLA with model-based rollouts and reward-based trajectory selection, inspired by Model Predictive Control algorithm. SITCOM leverages a learned dynamics model to simulate multi-step action rollouts to select the best candidate plan for real-world execution, transforming one-shot VLAs into robust long-horizon planners. We develop an efficient transformer-based dynamics model trained on large-scale BridgeV2 data and fine-tuned on SIMPLER environments to bridge the Real2Sim gap, and score candidate rollouts using rewards from simulator. Through comprehensive evaluation across multiple tasks and settings in the SIMPLER environment, we demonstrate that SITCOM when combined with a good reward function can significantly improve task completion rate from 48\% to 72\% using trained dynamics model. 
\end{abstract}

\section{Introduction}
Robot learning has long been constrained by the need for extensive, labeled data to train effective control policies \cite{mccarthy2024generalistrobotlearninginternet}. Traditional approaches often require meticulously annotated datasets with precise action labels, which are costly and time-consuming to acquire, especially for complex robotic tasks. Furthermore, reliance on task-specific datasets limits the generalization of learned policies to unseen environments, embodiments, or variations of the original task. Although reinforcement learning (RL) offers a potential solution, it exhibits poor sample efficiency for real-world problems and often requires significant--and sometimes unsafe--interaction with the environment. World models \cite{ha_2018_1207631,hafner2022masteringataridiscreteworld}, which learn predictive representations of the environment to improve sample efficiency and enable risk-free exploration, have emerged as a promising alternative. However, they still struggle to accurately model the diversity and complexity of real-world scenarios, limiting their effectiveness in open-ended tasks.

Vision–Language–Action (VLA) models have recently achieved remarkable success in interpreting natural language instructions and generating corresponding single-step control commands for robotic systems \cite{kim2024openvla,brohan2023rt1,octomodelteam2024}. Despite these advances, deploying VLAs in real-world robotics remains fraught with challenges: they often lack mechanisms for lookahead, struggle to recover from compounding errors, and cannot plan over long horizons in dynamic environments. Such limitations manifest in failures during multi-step tasks like sequential object manipulation, assembly, or navigation in cluttered scenes.

\begin{figure*}[t] 
    \centering
\includegraphics[width= \textwidth]{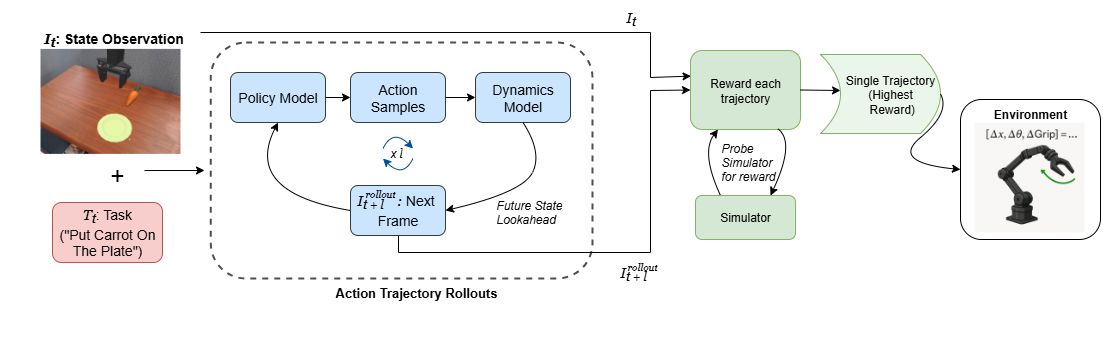}
    \caption{\footnotesize Method Diagram explaining the overall SITCOM architecture. Given an initial frame and goal in text description, a VLA predicts actions which is fed into the dynamics model to get the next frame. This process is then iteratively repeated with the output of dynamics model to generate action sequence rollouts. Finally, a reward model is used to rank the action sequences and pick the best one among them to execute in the real-world.}
    \label{fig:method}
    \vspace{-2mm}
\end{figure*}

In this work, we introduce the \emph{Scaling of Inference-Time Compute for VLAs} (\textbf{SITCOM}) framework (\figurename~\ref{fig:method}), which endows any pretrained VLA with model-based rollout and reward-ranking capabilities inspired by Model Predictive Control (MPC) \cite{mpc1999667}. At each decision step, SITCOM’s VLA policy proposes multiple candidate actions; a learned dynamics model then simulates resulting next states, and this process repeats to generate full multi-step trajectory rollouts. Finally, a reward model ranks these candidate sequences, selecting the trajectory with maximum reward for real-world execution. By performing this \emph{inference-time planning}, SITCOM transforms one-shot VLAs into robust long-horizon planners, improving both reliability and success rates in complex tasks.

In this paper,
\begin{itemize}
  \item We propose the SITCOM framework, a general-purpose inference-time planning framework that enhances any VLA by simulating multi-step action rollouts and selecting optimal action sequences through a reward mechanism.
  \item We introduce an efficient transformer-based dynamics model pre-trained on large-scale BridgeV2 ~\cite{walke2024bridgedatav2datasetrobot} data and fine-tuned on SIMPLER environment ~\cite{li24simpler} trajectories to mitigate the Real2Sim gap. To further address error accumulation during long-horizon rollouts, we introduce a DAgger-inspired~\cite{ross2011reduction} adaptation strategy that bridges the distributional shift between model inputs and predicted outputs.
    \item We provide insights on performance gains by rollouts of VLA over number action sequences and the depth of future predictions.
\end{itemize}
Finally, we validate our approach through comprehensive evaluations on multiple tasks and settings in the SIMPLER environment ~\cite{li24simpler}, demonstrating that SITCOM consistently outperforms strong baselines and enables robust long-horizon planning for robotic manipulation.

\section{Related Work and Background}
\textbf{Related Datasets}. Recent VLA research has drawn on a wide spectrum of datasets, from early human-teleoperated collections (MIME~\cite{sharma2018multiple}, RoboTurk~\cite{mandlekar2018roboturk}) and scripted large-scale efforts (RoboNet~\cite{dasari2019robonet}, MT-Opt~\cite{kalashnikov2021mt}) to more recent real-world corpora (BC-Z~\cite{jang2022bc}, RT-1-Kitchen~\cite{brohan2022rt}, OXE~\cite{vuong2023open}, DROID~\cite{khazatsky2024droid}). These resources span hundreds of tasks across diverse robots but differ widely in annotation granularity (raw controls vs. language), complicating unified VLA training. In this work, we build on BridgeData V2~\cite{walke2024bridgedatav2datasetrobot} as our primary pretraining corpus: 60k episodes of human and scripted trajectories on a WidowX-250 arm, covering 13 manipulation skills across 24 scenes with over 100 objects, paired with RGB–D, segmentation, and language goal annotations. Its scale, diversity, and grounding make it an ideal foundation for pretraining our Transformer-based dynamics model and fine-tuning OpenVLA within the SITCOM framework.  
\textbf{Vision-language models for robot generalization}. Recent advances in vision–language models (VLMs) have expanded robotic generalization by providing visuo-linguistic representations~\cite{radford2021learning, liu2024grounding}, image generation~\cite{rombach2022high}, and multimodal reasoning~\cite{liu2024visual, li2023blip, fried-pragmatics-2023}, enabling applications such as goal generation~\cite{black2023zero}, reward shaping~\cite{du2023vision, ma2023liv}, and representation learning~\cite{nair2022r3m, karamcheti2023language}. Vision–language–action models (VLAs)~\cite{brohan2023rt2, embodimentcollaboration2024openxembodimentroboticlearning, kim2024openvla} build directly on these advances, achieving state-of-the-art generalist control and strong transfer to novel objects and scenes, though often with limited use of the reasoning capabilities of pre-trained VLMs. Recent work has sought to address this via Chain-of-Thought (CoT) supervision~\cite{zawalski2024robotic, blank2024scaling}, explicitly teaching task decomposition through curated datasets. In contrast, SITCOM takes a complementary path: rather than explicit step-by-step reasoning, we leverage simulator-based rollouts and reward-driven selection to implicitly evaluate futures, guiding long-horizon behavior without requiring curated decompositions.

\textbf{World models for planning and control}. Predictive models of environment dynamics have long underpinned robotics and reinforcement learning~\cite{sutton1991dyna, holkar2010overview, williams2017information}. Recent work shows that forecasting future states in pixels~\cite{finn2017deep, ebert2018visual, ko2023learning, du2023learning} or latents~\cite{micheli2022transformers, hafner2019learning} can improve sample efficiency and planning: pixel models offer strong visual grounding but are computationally heavy, while latent models are efficient but often task-specific~\cite{hafner2023mastering, hansen2023td}. Large-scale generative video models~\cite{hu2023gaia, yang2023learning, bruce2024geniegenerativeinteractiveenvironments} push realism further but rely on expensive diffusion backbones and language prompts, limiting their use for fine-grained control. In SITCOM, we take a middle ground: an efficient transformer-based dynamics model that predicts future frames in pixel space, lightweight enough for multi-step rollouts. To ensure long-horizon reliability, we adapt it with a DAgger-style~\cite{ross2011reduction} strategy that fine-tunes on its own predictions, reducing distributional drift. Pretraining on BridgeV2 and adapting to SIMPLER tasks yields robust, scalable inference-time planning without costly diffusion architectures or dense task supervision.

\textbf{Reward models for robotic planning}. Recent work on reward modeling leverages large pre-trained models to evaluate outputs, with the “LLM-as-a-Judge” paradigm~\cite{gu2024survey} inspiring implementations such as JudgeLM~\cite{zhu2023judgelm} and Generative Judge~\cite{li2023generative}. While these efforts largely target language evaluation, robotics has adapted similar ideas to visual domains, using language models for reward design~\cite{yu2023language} or preference tuning~\cite{liu2025inference}. In SITCOM, we instead probe future simulator states to compute rewards: though unrealistic in real-world deployment, this strategy yields interpretable signals and highlights common success and failure modes of VLAs, enabling effective rollout selection during planning.
\section{Method}
\textbf{Notation:} Let $\mathcal{I}$ denote image observations, $\mathcal{T}$ is the task instructions, $\pi_\text{VLA}(\mathcal{I}, \mathcal{T})$ is the VLA model that outputs an action $a$, $f_\text{dyn}(\mathcal{I}, a)$ is the dynamics model, $r(\mathcal{I}_0, \mathcal{I}_l, \mathcal{T})$ is the reward model, $n$ is the number of trajectories, and $l$ is the rollout length.


We propose an inference algorithm for decision-making in an environment using Vision-Language-Action (VLA) models. 
At each inference step, the agent is provided with an observation $\mathcal{I}$ (an image) and a task instruction $\mathcal{T}$. 
The VLA model takes $(\mathcal{I}, \mathcal{T})$ as input and outputs an action. 
To encourage exploration, we set a high sampling temperature and sample $n$ candidate actions.

Each sampled action initializes a trajectory. 
A learned dynamics model, distinct from the real environment, is used to perform rollouts: 
given an image $\mathcal{I}$ and an action $a$, the dynamics model predicts the next image $\mathcal{I}'$.
Subsequent actions are also temperature-sampeld from the VLA model.

Each trajectory is simulated for $l$ steps and we do this for $k$ trajectories. To perform this simulation, we propose two methods - 1) SITCOM (EnvSim) that uses another instance of the environment to perform the rollouts, 2) SITCOM (Dynamics) where we used a trained dynamics model for the rollouts. 

After simulation, we give reward based on the final state of the trajectories (at the pre-defined depth). Our reward design incorporates the gripper-object gap, object-destination distance, and grasp success indicators. 
The trajectory with the highest reward score is selected, and the trajectory is executed in the real environment.
This procedure is repeated at the environment's replanning frequency until task success or termination. Algorithm \ref{algo:sitcom} demonstrates the entire pipeline. Details regarding architecture of dynamics model can be found in section \ref{sec:dynamics}.

\subsection{Training Dynamics Model}
\label{sec:dynamics}

\begin{figure*}[t]
  \centering
  \begin{minipage}[t]{0.49\textwidth}
    \centering
    \includegraphics[width=\linewidth]{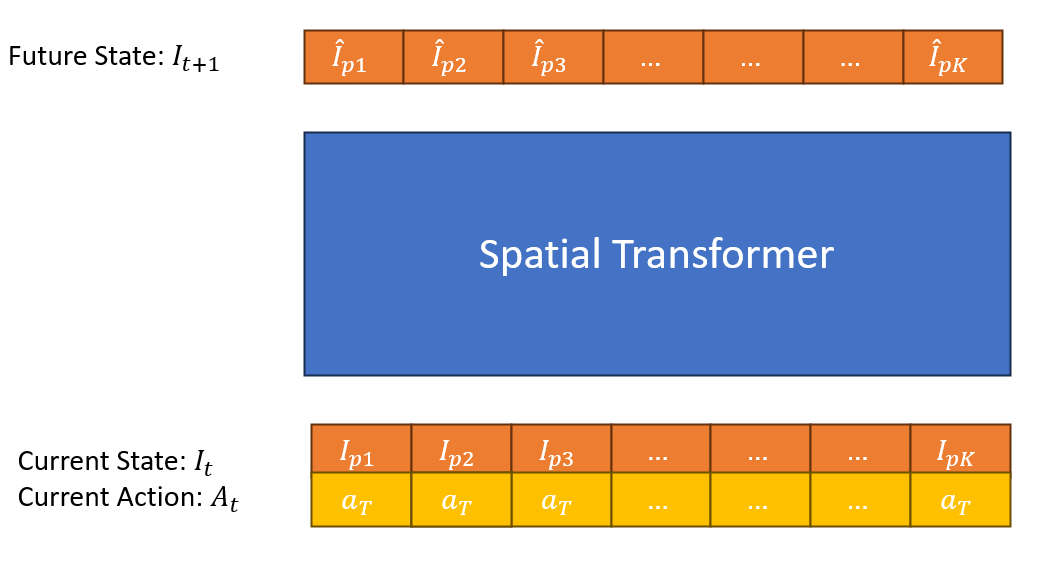}
    \caption{\footnotesize Dynamics Model Architecture}
    \label{fig:dynamics}
  \end{minipage}
  \hfill
  \begin{minipage}[t]{0.49\textwidth}
    \centering
    \includegraphics[width=\linewidth]{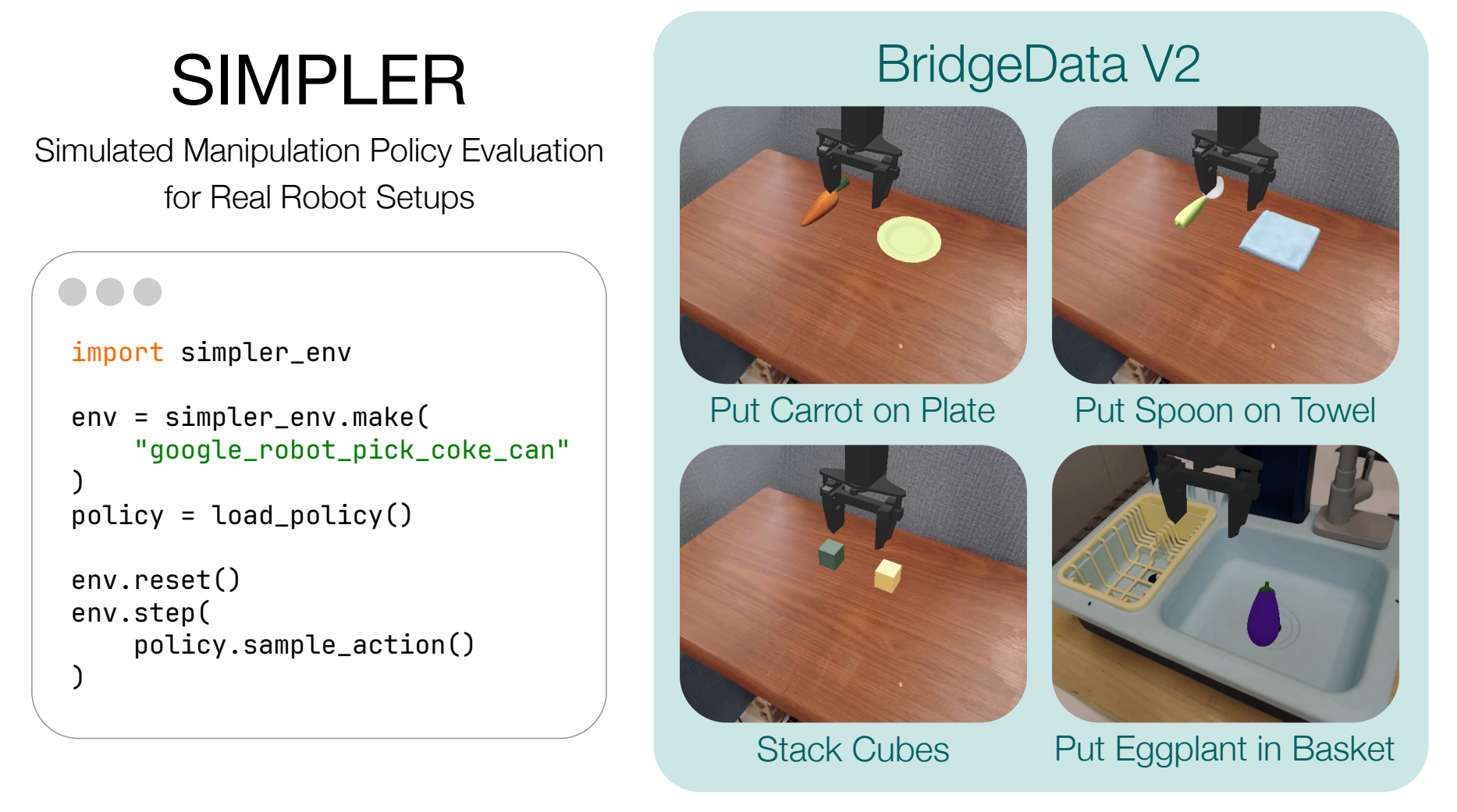}
    \caption{\footnotesize SIMPLER environment and four different tasks with WidowX arm}
    \label{fig:simpl}
  \end{minipage}
\end{figure*}

We train a dynamics model $f_{dyn}(.)$ that predicts the next state given the current state and action. The model uses an encoder-decoder architecture: the encoder processes image patches, concatenates them with action information, and the decoder predicts patches for the next state (\figurename~\ref{fig:dynamics}). To match inference conditions where the model rolls out autoregressively for $l$ steps, we train it to predict future states from its own predictions for $l_{train}$ steps (detailed in Section \ref{sec:analysis}).

The model is trained using combined L1 pixel-wise loss and Learned Perceptual Image Patch Similarity (LPIPS) loss. L1 loss ensures pixel-level accuracy, while LPIPS loss promotes perceptual realism by measuring similarity in deep feature space. This combination balances low-level precision with high-level visual coherence.

We leverage approximately 25,000 trajectories from BridgeV2~\cite{walke2024bridgedatav2datasetrobot} for pretraining, providing diverse manipulation scenarios that enable generalizable dynamics learning. To address the Real2Sim gap, we fine-tune on in-domain SIMPLER trajectories, adapting the model to the specific visual appearances and physics of our evaluation environment. This two-stage training approach improves robustness for long-horizon rollouts in the SITCOM framework.
\begin{algorithm}[t]
\footnotesize
\label{algo:sitcom}
\caption{Inference with VLA and Dynamics Model}
\SetAlgoLined
\KwIn{Initial observation $\mathcal{I}_0$, task instruction $\mathcal{T}$}
\While{task not completed and not terminated}{
    Sample $n$ actions $\{a_0^1, \dotsc, a_0^n\}$ from $\pi_\text{VLA}(\mathcal{I}_0, \mathcal{T})$ with high temperature\;
    \ForEach{$i \in \{1, \dotsc, n\}$}{
        Initialize trajectory with $\mathcal{I}_0^i \leftarrow \mathcal{I}_0$\;
        Apply action $a_0^i$ using dynamics model: $\mathcal{I}_1^i \leftarrow f_\text{dyn}(\mathcal{I}_0^i, a_0^i)$\;
        \For{$t = 1$ \KwTo $l-1$}{
            $a_t^i \leftarrow \arg\max_{a} \pi_\text{VLA}(\mathcal{I}_t^i, \mathcal{T})$\;
            $\mathcal{I}_{t+1}^i \leftarrow f_\text{dyn}(\mathcal{I}_t^i, a_t^i)$\;
        }
    }
    Compute rewards $r^i = r(\mathcal{I}_0, \mathcal{I}_l^i, \mathcal{T})$ for each $i$\;
    Select trajectory $i^* = \arg\max_i r^i$\;
    Execute trajectory $\{a_0^{i^*}, a_1^{i^*}, a_2^{i^*}, \dots a_{l-1}^{i^*} \}$ in the real environment\;
    Observe next environment state $\mathcal{I}_0$\;
}
\end{algorithm}

\subsection{Finetuning VLA}
\label{sec:loss_data}
We train the vision-language-action (VLA) model using a standard cross-entropy loss over discretized action tokens. Since no publicly available expert trajectory data existed for the SIMPLER environment and pretrained real-world models performed poorly in simulation (as shown in the table), we curated our own dataset of 100 expert trajectories.
Our dataset curation process involved three steps: first, we ran a pre-trained model \cite{ye2024latent} to generate initial trajectories; second, we applied heuristic rules to identify successful executions; and finally, we used human filtering to ensure high-quality expert demonstrations. These curated trajectories provide the VLA model with robust examples of successful task execution, enabling it to learn effective mappings from vision-language inputs to low-level control outputs.





\section{Experiment}
\subsection{Task Setup}
To benchmark the effectiveness of our method, we use SIMPLER \cite{li24simpler}, a suite of open-source simulated environments designed to evaluate generalist robot manipulation policies in a scalable and reproducible manner. We evaluate our model on four tasks (Figure \ref{fig:simpl}) using the 7-DOF WidowX robotic arm. Since SIMPLER does not provide fine-tuning trajectories, we collect 100 multi-task trajectories using successful rollouts from an open source VLA model trained on BridgeData V2 \cite{walke2024bridgedatav2datasetrobot}, while ensuring diverse object orientations and positions.

We evaluate baseline and our framework across four tasks for WidowX arm and five tasks for Google robot. We focus on following metrics to access perfromance.

\textbf{Average Success Rate} is our main metric that serves as a performance metric for evaluating our models. This metric quantifies the fraction of tasks successfully completed by a policy across diverse scenarios, calculated as:
\begin{equation*}
    \text{Average Success} = \frac{\text{\# Successful Trials}}{\text{Total Trials}}
\end{equation*}

\textbf{Partial Success Rate} captures instances where the robot achieves part of the goal. For example, the robot could be grasping an object but failing to place it correctly. This metric is important because many robotic tasks involve sequential steps\textit{} and analyzing partial success helps identify failure points and areas for improvement.

\textbf{Time ($\propto$ Compute)} captures the compute required to take a single action. This becomes an important factor for our evaluation, since we propose to scale test-time compute through rollouts. The number of rollouts is limited by the computation budget.


\subsection{Baselines}
We use the following three leading pretrained architectures as our baselines: \textbf{OpenVLA} \cite{kim2024openvla}, a 7B-parameter transformer that extends large-scale VLMs to action prediction; \textbf{RT-1-X}, an extension of the Robotics Transformer framework scaled to the full Open-X dataset for improved generalization across manipulation skills \cite{brohan2023rt1}; and \textbf{Octo} \cite{octomodelteam2024}, a diffusion-policy transformer (97M parameters) designed for cross-environment transfer.

\begin{table}[t]
\centering
\scalebox{0.75}{
{\small
\begin{tabular}{l|cccc|cccc}
\toprule
& \multicolumn{4}{c|}{\textbf{Complete Success}} & \multicolumn{4}{c}{\textbf{Partial Success}} \\
\midrule
\multicolumn{9}{c}{\textbf{Bridge Tasks}} \\
\midrule
\textbf{Task} & \textbf{RT1-X} & \textbf{Octo-base} & \textbf{Octo-small} & \textbf{OpenVLA} & \textbf{RT1-X} & \textbf{Octo-base} & \textbf{Octo-small} & \textbf{OpenVLA} \\
\midrule
Put spoon on tablecloth & 0.042 & 0.111 & \textbf{0.486} & 0.000 & 0.125 & 0.375 & \textbf{0.833} & 0.083 \\
Put carrot on plate & 0.083 & \textbf{0.097} & \textbf{0.097} & 0.000 & 0.250 & \textbf{0.472} & 0.264 & 0.167 \\
Stack green block on yellow block & 0.000 & 0.000 & 0.014 & \textbf{0.042} & 0.125 & 0.333 & \textbf{0.347} & 0.125 \\
Put eggplant in basket & 0.000 & 0.417 & \textbf{0.556} & 0.000 & 0.000 & 0.639 & \textbf{0.861} & 0.042 \\
\bottomrule
\end{tabular}
}
}
\caption{\footnotesize Performance of multimodal VLA models on different tasks, showing both complete success and partial success rates. Best performance for each task and metric is highlighted in \textbf{bold}.}
\label{tab:multimodal_results}
\end{table}

We benchmark multiple vision-language-action models in the SIMPLER~\cite{li24simpler} simulation environment (\tablename~\ref{tab:multimodal_results}) and identify the Real2Sim gap as a key bottleneck limiting performance. This gap arises from discrepancies between real-world inputs and simulated environments, leading to degraded transferability of policies. To study this systematically, we focus on OpenVLA~\cite{kim2024openvla} as our primary baseline, due to its popularity, open-source availability, and comprehensive documentation. To mitigate the Real2Sim gap, we fine-tune OpenVLA on in-domain simulated trajectories, improving its adaptation to the SIMPLER environment.

Initially, we evaluate our approach using an \emph{oracle simulator}, leveraging direct access to ground-truth environment states to benchmark improvements from fine-tuning and dynamics-based rollouts in a controlled setting. Building on this foundation, we then integrate our trained dynamics model and memory-based general reward model, enabling simulation of multi-step rollouts and scoring of candidate trajectories entirely from learned models. This transition allows us to move beyond oracle supervision toward a fully self-contained, scalable simulator framework, setting the stage for broader deployment across diverse robotic tasks.

\section{Results and Analysis}
\label{sec:analysis}

All reported success rates are computed using reward signals obtained from the simulator (hence assuming perfect knowledge of the environment). We also ablate the effect of predicting future states using either the learned dynamics model or the oracle simulator to isolate sources of error.

\subsection{Overall Performance Comparison}
We first present the overall performance of our SITCOM framework compared to baseline methods in Table~\ref{tab:main_results}. For our main experiments, we use the following SITCOM configuration: rollout length of 10 steps and 5 candidate rollouts. The results demonstrate that SITCOM (EnvSim) achieves the highest performance across all tasks, with SITCOM (World Model) performing comparably. Both variants significantly outperform the baseline methods, OpenVLA and OpenVLA-SFT.

\begin{table}[t]
\centering
\scalebox{0.8}{
\begin{tabular}{@{}lcccc@{}}
\toprule
Tasks & OpenVLA & OpenVLA-SFT & SITCOM (EnvSim) & SITCOM (World Model) \\
\midrule
Put Carrot on Plate & 0.0 & 0.50 & 0.71 & 0.66 \\
Put Spoon on Table Cloth & 0.0 & 0.63 & 0.83 & 0.83 \\
Stack Green Block on Yellow Block & 0.042 & 0.17 & 0.58 & 0.62 \\
Put Eggplant in the Basket & 0.0 & 0.63 & 0.92 & 0.79 \\
\midrule
Overall (Avg) & 0.01 & 0.48 & \textbf{0.76} & 0.72 \\
\bottomrule
\end{tabular}
}
\caption{\footnotesize Success rates for different methods across robotic manipulation tasks. SITCOM results shown for configuration with rollout length=10, candidates=5}
\label{tab:main_results}
\footnotesize
   \centering
   \begin{minipage}{0.49\textwidth}
       \centering
       \begin{tabular}{|l|c|c|}
           \hline
           \textbf{Model}           & \textbf{FID (↓)} & \textbf{OFL (↓)} \\ 
           \hline
           BM. (BridgeV2)     & 17.0              & 1.665                        \\  
           FT. Model          & \textbf{11.2}     & \textbf{0.992}                  \\  
           \hline
       \end{tabular}
       \caption{\footnotesize Comparison of base model (BM) and fine-tuned (FT) model on FID and OFL metrics on transitions extracted from SIMPLER environment. Fine-tuning on in-domain SIMPLER trajectories improves both metrics, demonstrating better prediction of realistic and temporally coherent future state images.}
       \label{tab:dm_results}
   \end{minipage}
   \hfill
   \begin{minipage}{0.49\textwidth}
        \footnotesize
        \centering
        \begin{tabular}{@{}lccccccc@{}}
        \toprule
        Candidates & 1 & 5 & 10 & 15 & 20 & 25 \\
        \midrule
        Time (s)   & 21 & 35 & 75 & 100 & 130 & 160 \\
        \bottomrule
        \end{tabular}
        \caption{\footnotesize Planning time (in seconds) for different numbers of candidates}
        \label{tab:planning_time}
   \end{minipage}
\end{table}

As shown in Table~\ref{tab:main_results}, SITCOM (EnvSim) achieves an average success rate of 76\%, while SITCOM (World Model) achieves 72\%, both dramatically outperforming the baseline OpenVLA (1\%) and OpenVLA-SFT (48\%) models. This demonstrates the effectiveness of our simulation-guided approach for improving robotic manipulation performance. In the following sections, we analyze how varying these configuration parameters affects performance. Next, we perform various experiments for both SITCOM-EnvSim and SITCOM-Dynamics model, and also conduct ablation studies for the dynamics and reward models.

\subsection{Analysis of rollout parameters} 

\begin{figure*}[t]
    \centering
    \begin{minipage}[t]{0.48\textwidth}
        \centering
        \includegraphics[width=\linewidth]{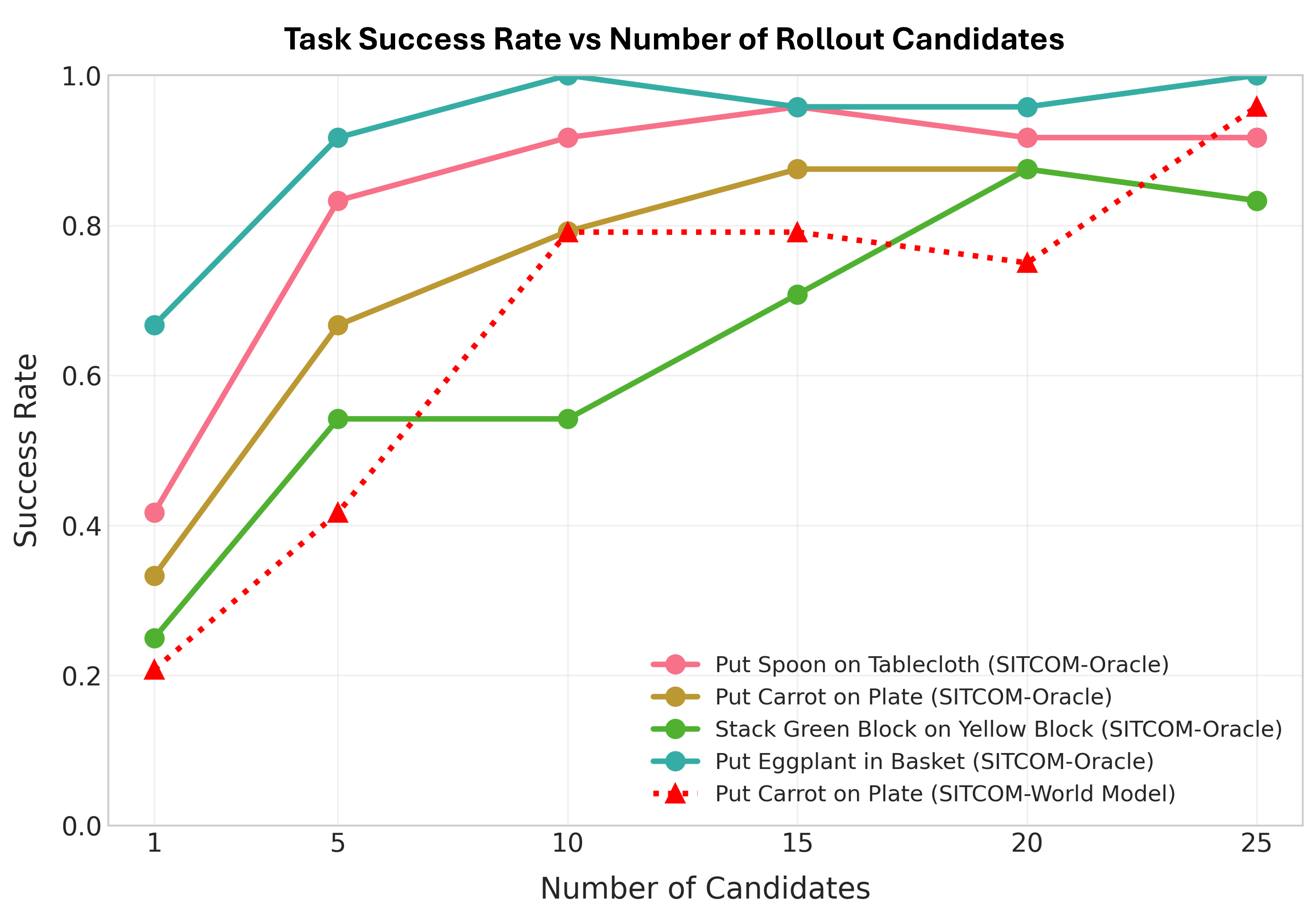}
        \caption{\footnotesize Scaling with number of rollouts for SITCOM-EnvSim and SITCOM-Dynamics models.}
        \label{fig:vary_candidates}
    \end{minipage}
    \hfill
    \begin{minipage}[t]{0.48\textwidth}
        \centering
        \includegraphics[width=\linewidth, height=0.69\linewidth]{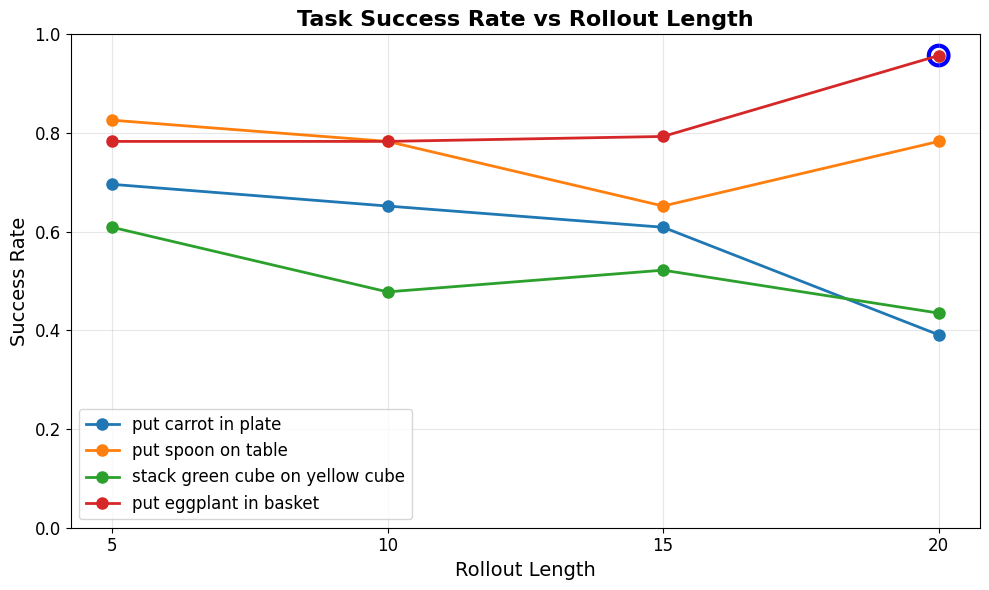}
        \caption{\footnotesize Scaling with varying rollout lengths. The performance of the dynamics model decreases with the increase in rollout length, explaining the drop in overall performance for some of the tasks.}
        \label{fig:vary_rollouts}
    \end{minipage}
\end{figure*}

\textbf{Breadth: Number of rollout candidates}. First, we demonstrate that scaling the number of rollout candidates is beneficial for robotic tasks. We observe continuous gains until 25 candidates for some tasks, while for other tasks the gains begin to saturate before, as shown in Figure~\ref{fig:vary_candidates}. 
We also report the time taken for varying candidates in Table~\ref{tab:planning_time}.




While we notice that time increases with scaling the number of candidates, the entire pipeline supports parallelism, allowing this time to be controlled. We elaborate on this in the future work section.

\textbf{Depth: Length of Rollouts}. Next, we vary the rollout lengths for the SITCOM-Dynamics model. The optimal rollout length may vary across tasks, as some tasks require more in-depth planning while others benefit from short-term planning. While we plan to develop a method to estimate the optimal rollout length for each task, we currently keep it fixed and experiment with various rollout lengths.

We find that for challenging tasks such as "Put Eggplant in Basket," our model benefits from longer rollouts, as demonstrated in Figure~\ref{fig:vary_rollouts}. The results show that performance improves as we increase the rollout length, with particular gains observed in complex manipulation tasks.

\subsection{Dynamics Model}

\begin{figure}[t]
    \centering
    \includegraphics[width=\textwidth]{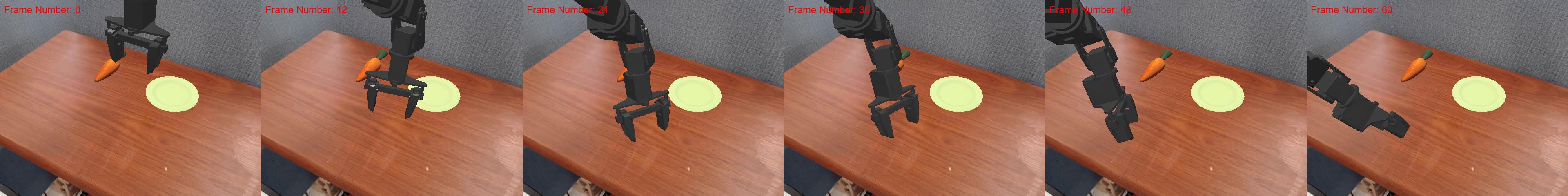}
    \caption{\footnotesize Failure case: The robot arm failed to reach the source object \textit{carrot} using OpenVLA base.}
    \label{fig:failure_carrot}
\end{figure}

To test the effectiveness of our dynamics model, we evaluate it using Frechet Inception Distance (FID) and Optical Flow Loss (OFL) scores. Frechet Inception Distance (FID) measures how closely predicted future state images resemble ground-truth images, with lower scores indicating more realistic predictions. Optical Flow Loss (OFL) evaluates how well the dynamics model captures temporal changes by comparing pixel-wise motion between consecutive frames, specifically designed to assess whether the model predicts meaningful task-relevant dynamics rather than copying static background information. 

Table \ref{tab:dm_results} compares the base model (trained only on the BridgeV2 dataset) and the fine-tuned model (fine-tuned on in-domain SIMPLER trajectories) using the above metrics.


The results in Table \ref{tab:dm_results} highlight both the effectiveness of our base dynamics model and the benefits of fine-tuning. Even without fine-tuning, the Base Model achieves a respectable FID of 17.0, demonstrating that training on large-scale BridgeV2 trajectories allows the model to generalize reasonably well to unseen environments. This baseline performance highlights the generalizability of our model to unseen environments, underscoring the effectiveness of training on diverse, large-scale datasets.

Fine-tuning on in-domain SIMPLER trajectories further improves performance, lowering the FID from 17.0 to 11.2 and reducing the OFL from 1.665 to 0.992. This shows that fine-tuning enhances the model’s ability to capture task-relevant temporal dynamics while minimizing redundant static information. These improvements help address the Real2Sim gap by adapting the model to environment-specific dynamics, leading to more accurate and temporally coherent predictions.

Overall, these results demonstrate that our world model provides a solid foundation for dynamics prediction even before fine-tuning, and that the additional fine-tuning step further improves the model’s robustness, temporal coherence, and adaptability to the target simulation environment, thereby helping bridge the Real2Sim gap.

\section{Qualitative Discussion}

We analyze the components of SITCOM with a focus on policy behavior and failure modes.

\textbf{VLA Policy.} As shown in \tablename~\ref{tab:main_results}, the baseline OpenVLA~\cite{kim2024openvla} achieves a partial success rate of 0.167 but no full completions on the \texttt{PutCarrotOnPlate} task, while fine-tuning on SIMPLER trajectories raises the success rate to 0.500, reducing the Real2Sim gap.

\textbf{Failure Modes.} OpenVLA exhibits two main limitations: (i) the \textbf{Real2Sim gap}, where differences in dynamics and visual affordances hinder transfer from real data to simulation, and (ii) \textbf{limited generalization}, as imitation learning struggles with out-of-distribution scenes. These failures are especially pronounced in fine-grained control tasks such as \texttt{PutCarrotOnPlate}, where the agent must grasp at the right affordance and close the gripper precisely; small deviations cause the carrot to slip, as illustrated in Figure~\ref{fig:failure_carrot}.


\textbf{Finetuning OpenVLA to Reduce Real2Sim GAP}. We create this model by fine-tuning OpenVLA on $\sim100$ 
trajectories from SIMPLER. The goal is to bridge the Real2Sim gap present in the base OpenVLA. We observe a performance boost of approximately 40\% after fine-tuning. The model is now better calibrated for the simulator setting; however, its performance is still around 40\%, and we aim for further improvements.

\textbf{Leveraging Inference-Time Compute Using Simulators for Rollouts}.
We scale our model using test-time compute through a guidance mechanism that employs an oracle simulator for rollouts. Our reward design incorporates the gripper-object gap, object-destination distance, and grasp success indicators, generalizing well across Bridge dataset tasks. While these rewards may not generalize to all scenarios, they extend to broader robotic manipulation tasks. This approach achieves 80\% success on Bridge tasks, significantly outperforming previous methods through effective policy sampling and reward design. Although action selection time increases, the pipeline parallelizes across multiple GPUs to mitigate computational overhead.


\begin{figure}[t]
    \centering
    \includegraphics[width=\textwidth]{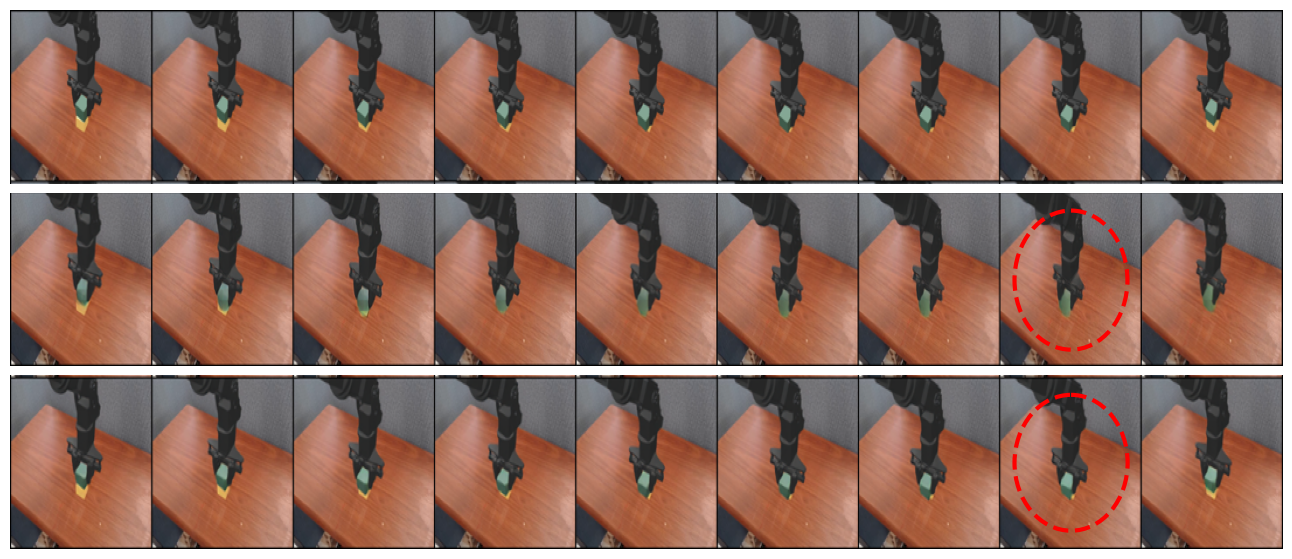}
    \caption{\footnotesize \textbf{Qualitative comparison of world model rollouts.} Top row: Ground truth trajectory. Middle row: Rollouts from a world model trained without DAgger-style adaptation. Bottom row: Rollouts from a world model trained with DAgger-style adaptation. Object reconstruction issues without adaptation are highlighted.}
    \label{fig:traj_comparison}
\end{figure}

\begin{figure}[t]
    \centering
    \includegraphics[width=\textwidth]{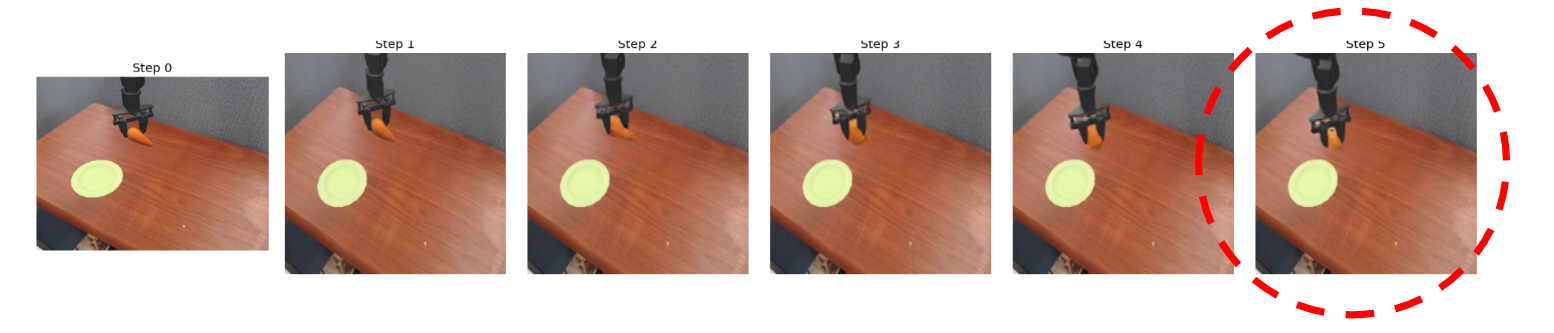}
    \caption{\footnotesize \textbf{Example of poor object reconstruction during world model rollout.} Despite improvements from DAgger-style adaptation, occasional failures in object consistency remain evident at longer rollout horizons.}
    \label{fig:poor_wm_reconstruction}
\end{figure}

\textbf{World Model}. Initially training our world model to predict single timesteps ahead resulted in significant object reconstruction errors during extended rollouts due to compounding prediction errors (Figure~\ref{fig:traj_comparison}, second row). We addressed this using DAgger-style adaptation, where the model uses its own predictions as inputs during training, markedly improving reconstruction over longer horizons (third row). However, prediction drift and reconstruction failures persist during test-time (Figure~\ref{fig:poor_wm_reconstruction}), suggesting policy-generated trajectories remain out-of-distribution for our world model. We attribute this to training exclusively on successful trajectories and hypothesize that incorporating unsuccessful trajectories would mitigate these errors.

\section{Future work and Limitations}
While SITCOM demonstrates promising results in simulation, several limitations remain, providing opportunities for future research.
\begin{enumerate}[nosep]
    \item \textbf{Limited Exposure to Failure States}. To address model bias from training solely on successful trajectories, we will augment our dataset with collected failure examples to improve robustness and accurate penalization in off-distribution states.
    \item \textbf{Real-World Deployment Challenges}. To bridge the sim2real gap, future work will focus on addressing challenges like visual and physical discrepancies through techniques such as real-world fine-tuning, domain adaptation for vision, and closed-loop replanning to correct accumulated errors during execution.
    \item \textbf{Deterministic Dynamics Model Limitation}. To overcome the limitations of our deterministic dynamics model in handling stochastic environments, we propose exploring probabilistic, action-conditioned video diffusion models. This approach would better capture uncertainty and enable more flexible and robust planning for complex, real-world manipulation tasks by generating diverse and plausible future outcomes.
    \item \textbf{Inference-Time Bottleneck and Control Frequency}. To overcome the inference-time bottleneck that currently limits real-world control frequency, future work will focus on reducing latency by parallelizing rollout generation and exploring action chunking. Additionally, we propose using action-conditioned video diffusion models to simulate entire trajectories in a single pass, significantly improving computational efficiency for high-frequency, dexterous manipulation tasks.
\end{enumerate}

\bibliographystyle{unsrtnat}
\bibliography{references}


\end{document}